# An attempt to generate new bridge types from latent space of generative flow

Hongjun Zhang

Wanshi Antecedence Digital Intelligence Traffic Technology Co., Ltd, Nanjing, 210016, China

583304953@QQ.com

**Abstract:** Through examples of coordinate and probability transformation between different distributions, the basic principle of normalizing flow is introduced in a simple and concise manner. From the perspective of "the distribution of random variable function", the essence of probability transformation is explained, and the scaling factor Jacobian determinant of probability transformation is introduced. Treating the dataset as a sample from the population, obtaining normalizing flow is essentially through sampling surveys to statistically infer the numerical features of the population, and then the loss function is established by using the maximum likelihood estimation method. This article introduces how normalizing flow cleverly solves the two major application challenges of high-dimensional matrix determinant calculation and neural network reversible transformation. Using symmetric structured image dataset of three-span beam bridge, arch bridge, cable-stayed bridge and suspension bridge, constructing and training normalizing flow based on the Glow API in the TensorFlow Probability library. The model can smoothly transform the complex distribution of the bridge dataset into a standard normal distribution, and from the obtained latent space sampling, it can generate new bridge types that are different from the training dataset.

## 0  Introduction

There are two main categories of bridge-type innovation: ① Innovation in structural mechanics form. Create new bridge types that are different from the four basic structural forms of beam bridge, arch bridge, cable-stayed bridge, and suspension bridge, in order to achieve breakthroughs in span, reduce material consumption, and increase bearing capacity. To achieve this goal, in addition to breakthroughs in structural theory, it is also subject to external conditions such as materials, calculation methods, and construction techniques; ② Innovation in appearance design. The vast majority of bridges only serve the purpose of crossing obstacles, which is plain and without any form requirements. However, a few bridges in certain cities, scenic spots, and special occasions carry aesthetic requirements. At this point, bridge designers need to organically combine structure and artistic form, achieving a unity of function and form.

Throughout history, bridge-type innovation has always been a major challenge faced by bridge designers, and it is a challenging and attractive task. Generative artificial intelligence will be a powerful tool for bridge engineers to enhance their own abilities.

The Variational Autoencoder (VAE), Generative Adversarial Network (GAN), and Pixel Convolutional Neural Network (PixelCNN) used by the author in previous papers[1-3] represent different approaches to solve the problem of modeling sample space p(x) of image. The sample space of image is an extremely complex high-dimensional space with an astronomical space capacity, and only a few samples have practical significance. Direct modeling is difficult to operate. VAE and GAN compress the sample space into a low dimensional and dimensionally independent latent space. PixelCNN maintains the number of dimensions in the sample space unchanged, captures the dependency relationship between adjacent dimensions, and approximates the joint probability distribution through conditional probability product.

Normalizing flow (NF)[4-7] also maintains the number of dimensions in the sample space unchanged (similar to pixelCNN). It constructs a reversible transformation to transform the complex joint probability distribution p(x) in the sample space into a simple and universal joint probability distribution p(z) in the latent space. Then, z is sampled from the latent space and the sample space x is generated through an inverse transformation.

This article adopts the Glow (Generative Flow), based on the same dataset as before, and further attempts to bridge innovation (open source address of this article's dataset and source code: https://github.com/QQ583304953/Bridge-Flow).

# 1 Introduction to normalizing flow

## 1.1 Overview

1. VAE and GAN only implicitly learn the distribution in the data space and do not provide an analytical form of probability distribution function. Autoregressive models and normalizing flow have clear probability distribution functions. However, the autoregressive model has obvious drawbacks. Firstly, it needs to be generated point by point, which makes parallel computing difficult. Moreover, the generation of each variable only depends on a part of the other variables, and the accuracy is inherently insufficient. Normalizing flow does not have such drawbacks.

Normalizing flow is a reversible process that non-destructively transforms any complex data distribution into a simple basic distribution (using standard normal distribution here), and vice versa. Using the analytical formula of the known high-dimensional standard normal distribution probability density function, invert the joint probability distribution in the sample space. Its conception is clever and exquisite. Normalizing flow essentially uses sampling surveys to statistically infer the numerical features of the population, and the inferred results depends mainly on the quality of the dataset.

There are several architectures for normalizing flow, including NICE, RealNVP, Glow, FFJORD, etc.[4-7].

2. The dimensions of latent space of the normalizing flow are independent of each other, and the number of spatial dimensions equals the number of pixels. During sampling, a series of layer calculations are used to transform the latent space coordinates into the sample space coordinates, thereby generating meaningful images.

Although dimension independence is similar to Naive Bayes algorithm, Naive Bayes algorithm lacks distribution transformation steps, so normalizing flow is fundamentally different from Naive Bayes algorithm.

## 1.2 Examples of distribution transformation

In theory, as long as it is constantly transformed, the standard normal distribution can fit any complex distribution, and any complex distribution can also be transformed into a standard normal distribution. The transformation example is as follows:

1. Firstly, generate 5000 sample points that follow two-dimensional standard normal distribution. This moment, the two dimensions are independent of each other.

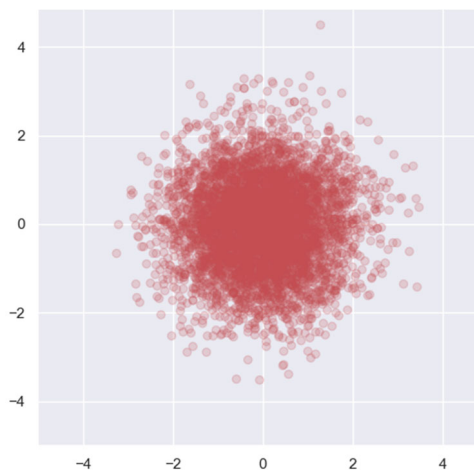

Fig.1 5000 sample points that follow two-dimensional standard normal distribution

2. Then perform a stretching transformation of the horizontal axis, and the transformation matrix

is $\begin{pmatrix} 10 & 0 \\ 0 & 1 \end{pmatrix}$. The two dimensions are still independent of each other.

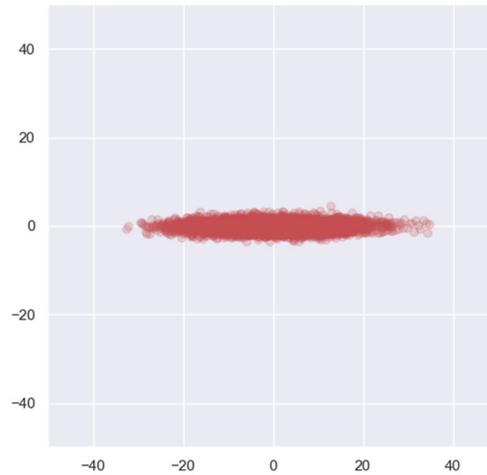

Fig.2 The result of stretching transformation of horizontal axis

3. Next, perform a rotation transformation, and the transformation matrix is $\begin{pmatrix} cos75° & -sin75° \\ sin75° & cos75° \end{pmatrix}$. This moment, the two dimensions are no longer independent and have a linear relationship. This is because after the rotation transformation, each dimension contains information from another dimension (such as x'=xcos75°-ysin75°).

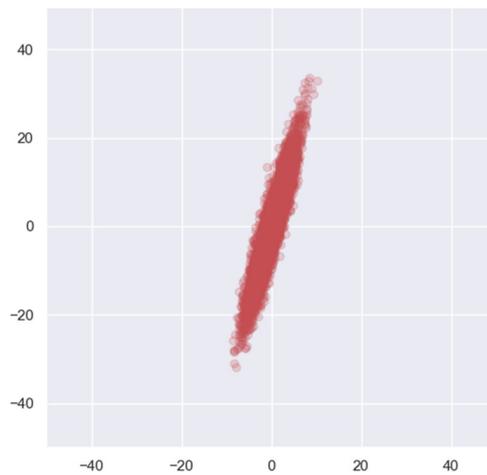

Fig.3 The result of rotation transformation

4. Next, perform a nonlinear shear transformation, and the transformation matrix is $\begin{pmatrix} 1 & x_i/40 \\ 0 & 1 \end{pmatrix}$. This moment, there is a nonlinear relationship between the two dimensions.

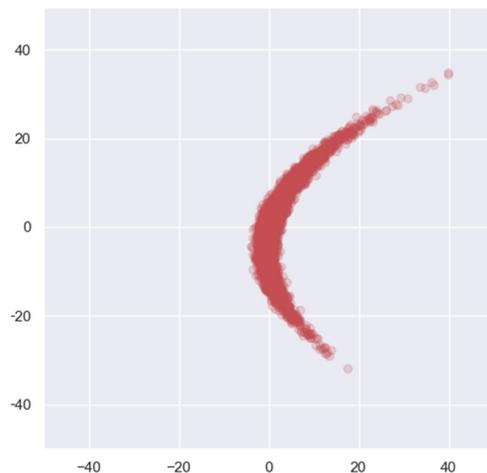

Fig.4 The result of nonlinear shear transformation

It can be seen that after the steps 1 to 4 above, the standard normal distribution is converted into a complex distribution.

5. The complex distribution in the above figure can also be converted into a standard normal distribution.

First, perform a nonlinear shear transformation, and the transformation matrix is the inverse matrix of $\begin{pmatrix} 1 & x_i/40 \\ 0 & 1 \end{pmatrix}$. This step eliminates the non-linear relationship of two dimensions and returns to the previous linear relationship.

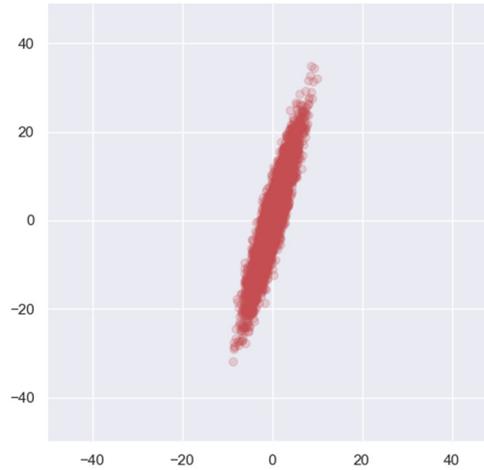

Fig.5 The result of inverse nonlinear shear transformation

6. Then the rotation transformation is performed, and the transformation matrix is the inverse matrix of $\begin{pmatrix} cos75° & -sin75° \\ sin75° & cos75° \end{pmatrix}$. This step eliminates the 2-dimensional linear relationship and restores the previous independent relationship.

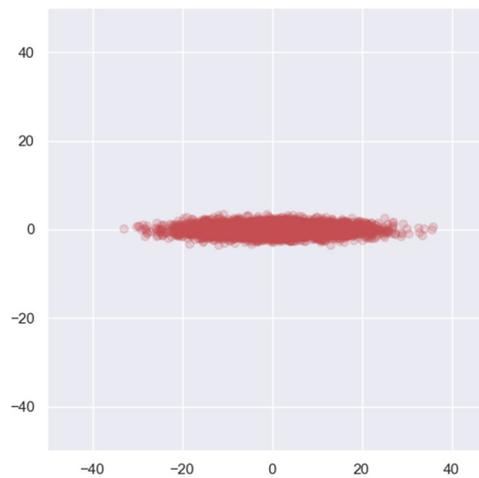

Fig.6 The result of inverse rotation transformation

7. Then perform a stretching transformation of the horizontal axis,, the transformation matrix is the inverse matrix of $\begin{pmatrix} 10 & 0 \\ 0 & 1 \end{pmatrix}$, the result is the standard normal distribution.

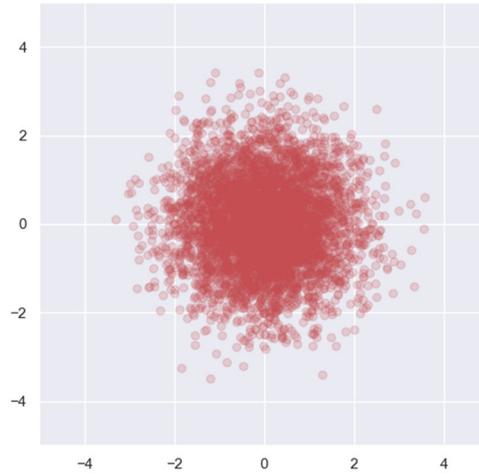

Fig.7 The result of inverse stretching transformation of horizontal axis

It can be seen that after the steps 5 to 7 above, a complex distribution is converted into a standard normal distribution.

### 1.3 Distribution of random variable function and Jacobian determinant

Normalizing flow maintains the number of dimensions in the sample space unchanged, so the solution and transformation of probability distribution is the key problem in modeling. And this key problem is essentially to solve the "distribution of random variable function". The following are analyzed by differential method, using straight lines instead of curves when undergoing nonlinear transformations.

1. First, give a one-dimensional example. Assuming a disc with a diameter D uniformly distributed on [5,6], i.e. p(D)=1, solve for the probability distribution density p(A) of the disc area A.

Solution: In order to ensure the same probability before and after the transformation, $p(D) * dD = p(A) * dA$, that is, $p(A) = p(D) * |\frac{dD}{dA}|$. If the logarithm is taken, it is $\ln p(A) = \ln p(D) + \ln|\frac{dD}{dA}|$.

$A = \pi/4 * D^2$, $A \in [25\pi/4, 9\pi]$, $D = |\pm\sqrt{4A/\pi}|$, $|\frac{dD}{dA}| = (\sqrt{4A/\pi})' = 1/\sqrt{A*\pi}$, so $p(A) = 1 * 1/\sqrt{A*\pi} = 1/\sqrt{A*\pi}$.

$|\frac{dA}{dD}| = p(D)/p(A)$, it can be considered that $|\frac{dA}{dD}|$ is a scaling factor for probability distribution density.

2. Taking two-dimensional as an example. Assuming a transformation is $\begin{pmatrix} y_1 \\ y_2 \end{pmatrix} = \begin{pmatrix} a & b \\ c & d \end{pmatrix} \begin{pmatrix} x_1 \\ x_2 \end{pmatrix} = \begin{cases} ax_1 + bx_2 \\ cx_1 + dx_2 \end{cases}$. In order to ensure the same probability before and after transformation, $p(y) * dA_y = p(x) * dA_x$, and the scaling factor of probability distribution density $|\frac{dA_y}{dA_x}| = p(x)/p(y)$.

According to the geometric meaning of the determinant, it can be inferred that the determinant $\det\begin{pmatrix} a & b \\ c & d \end{pmatrix}$ is the graph area scaling factor of a linear transformation. Therefore, $|\frac{dA_y}{dA_x}| = |\det\begin{pmatrix} a & b \\ c & d \end{pmatrix}|$.

So, $p(x) = p(y) * |\det\begin{pmatrix} a & b \\ c & d \end{pmatrix}|$.

And $\frac{\partial y_1}{\partial x_1} = \frac{\partial(ax_1 + bx_2)}{\partial x_1} = a$, Extend to vector to vector differentiation, Jacobian matrix: $J =$

$$\begin{pmatrix} \frac{\partial y_1}{\partial x_1} & \frac{\partial y_1}{\partial x_2} \\ \frac{\partial y_2}{\partial x_1} & \frac{\partial y_2}{\partial x_2} \end{pmatrix} = \begin{pmatrix} a & b \\ c & d \end{pmatrix}.$$

That is: p(x)=p(y)* |det$J$|. If the logarithm is taken, lnp(x)=lnp(y)+ln|det$J$|.

The same applies to the number of other dimensions.

### 1.4 Maximum likelihood estimation method

The basic idea of maximum likelihood estimation method is to use the concept of "the event with the highest probability is most likely to occur" to select appropriate population parameters $\hat{\theta}$. The parameters $\hat{\theta}$ maximizes the likelihood function L(θ) of the sample set, and take $\hat{\theta}$ as an estimate of the population parameter θ.

For example, if a coin with potentially uneven texture is thrown 4 times, and the result is 2 times for both the front and back facing upwards, what is the prior probability p of the coin facing upwards?

The rough analysis of this question is as follows: the prior probability p may be any value between 0 and 1, but the joint probability distribution L(p) of the result of this question varies depending on the value. For example, if p=0.4, then the joint probability distribution L(p=0.4)=p(x₁)* p(x₂)* p(x₃)* p(x₄)=$\frac{4}{10}*\frac{4}{10}*\frac{6}{10}*\frac{6}{10}=\frac{576}{10000}$; If p=0.5, then the result of this question is L(p=0.5)=$\frac{5}{10}*\frac{5}{10}*\frac{5}{10}*\frac{5}{10}=\frac{625}{10000}$; If p=0.6, then the result of this question is L(p=0.6)=$\frac{6}{10}*\frac{6}{10}*\frac{4}{10}*\frac{4}{10}=\frac{576}{10000}$. It is evident that p=0.5 is a more appropriate estimate for the population parameter. The exact solution procedure is omitted.

For the convenience of calculation, the likelihood function usually takes logarithmic values, lnL(θ)=$\sum_{i=1}^{n} lnp(x_i)$. When the population is a continuous random variable, $p(x_i)$ is the probability density.

### 1.5 Loss function

It is extremely difficult to directly model the joint probability distribution of the sample space for complex distributions. And we can write an analytical formula for the probability density function of a high-dimensional normal distribution, and use the distribution of the random variable function to calculate the joint probability distribution in the sample space.

Normalizing flow uses the method of distribution transformation to indirectly model the sample space. By calculating the probability density p(y) of the sample in the standard normal distribution space and the probability density scaling factor |det$J$| during neural network transformation, the probability density p(x) of the sample space can be indirectly calculated, that is, lnp(x)= lnp(y)+ln |det$J$|.

The training dataset is essentially sampling from the population, and the maximum likelihood estimation method is statistical inference of population parameters through sampling. So by maximizing the likelihood function lnL(θ), a reasonable estimate of the population parameter θ can be obtained. That is to say, Loss=-lnL(θ)=-$\sum_{i=1}^{n} lnp(x_i)$.

### 1.6 Challenges in practice

Normalizing flow faces two challenges in practice: ①the computational cost of computing high-dimensional matrix determinants is too high; ②Usual neural network calculations are unidirectional and cannot achieve lossless inverse transformations of input and output.

Here, taking the RealNVP model as an example[5,8], we demonstrate how to construct a neural network that transforms complex distributions into simple normal distributions, with determinant that is easy to solve. And the neural network can achieve lossless inverse transformations of input and output. The overall process is as follows:

1. During training: Firstly using mask, the coordinates of the dataset samples are divided into two parts $X_{1:d}$ and $X_{d+1:D}$. Let $X_{1:d}$ calculate the scaling factor s and translation factor t through the coupling layer, and use these two factors to perform coordinate transformation ($z=\frac{x-t}{s}$) on $X_{d+1:D}$ to obtain the latent space coordinate $Z_{d+1:D}$. The latent space coordinate $Z_{1:d}$ is directly equal to $X_{1:d}$. Taking the simplest two-dimensional case as an example, $\begin{pmatrix}z_1\\z_2\end{pmatrix} = \begin{cases}x_1\\-\frac{t}{s}+\frac{1}{s}*x_2\end{cases} = \begin{pmatrix}1 & 0\\c & \frac{1}{s}\end{pmatrix}\begin{pmatrix}x_1\\x_2\end{pmatrix}$, note this is the lower triangular matrix. At the same time, the coupling layer is stacked in pairs, so the mask mode is alternately used, so that the unchanged part of the coordinate is updated in the next layer.

The reason for such unconventional operations is to make the matrix of each coordinate transformation a lower triangular matrix, which makes the determinant equal to the product of diagonal elements, which is easy to calculate. The determinant of matrix product is the product of determinant, so the stacking of coupling layers does not increase the difficulty of determinant calculation, thus solving the first challenge.

2. When sampling: Firstly using mask, the latent space sampling coordinates are divided into two parts $Z_{1:d}$ and $Z_{d+1:D}$. Let $Z_{d+1:D}$ calculate the scaling factor s and translation factor t through the coupling layer (note during training, $Z_{d+1:D}=X_{d+1:D}$, and the weight parameters of the coupling layers are from the training stage, so the values of s and t here must be consistent with those during training). Use these two factors to perform coordinate transformation on $Z_{1:d}$ ($x=z*s+t$), and obtain the sample space coordinate $X_{1:d}$. And $X_{d+1:D}$ is directly equal to $Z_{d+1:D}$.

The mask mode is alternately used in multiple layers, then the unchanged part of the coordinate is updated in the next layer. The order of the layers here is the exact opposite of what it was in training. The inverse of the composition of two functions is the composition of its inverse, thus solving the second challenge.

## 1.7 Glow

Presented at NeurIPS 2018, Glow was one of the first models to demonstrate the ability of normalizing flows to generate high-quality samples and produce a meaningful latent space that can be traversed to manipulate samples. The key step was to replace the reverse masking setup with invertible 1 × 1 convolutional layers. For example, with RealNVP applied to images, the ordering of the channels is flipped after each step, to ensure that the network gets the chance to transform all of the input. In Glow a 1 × 1 convolution is applied instead, which effectively acts as a general method to produce any permutation of the channels that the model desires. The authors show that even with this addition, the distribution as a whole remains tractable, with determinants and inverses that are easy to compute at scale[8].

## 2 An attempt to generate new bridge types from latent space of Glow

### 2.1 Dataset

Using the dataset from the author's previous paper [1-3], which includes two subcategories for each type of bridge (namely equal cross-section beam bridge, V-shaped pier rigid frame beam bridge, top-bearing arch bridge, bottom-bearing arch bridge, harp cable-stayed bridge, fan cable-stayed bridge, vertical_sling suspension bridge, and diagonal_sling suspension bridge), and all are three spans (beam bridge is 80+140+80m, while other bridge types are 67+166+67m), and are structurally symmetrical.

For a 512x128 pixel image, the number of dimensions of latent space of Normalizing flow is 65536. Therefore, sufficient samples are needed to achieve better learning results. Two measures are taken here to increase sample quantity of the dataset: ① Using OpenCV to interpolate the images in the original dataset, thereby increasing the total number of images from 9600 to 108416; ② Using data augmentation, which involves adding Gaussian noise to each image. (Note: These two measures cure the symptoms rather than the root cause. Sampling is better only by increasing the types of bridge.)

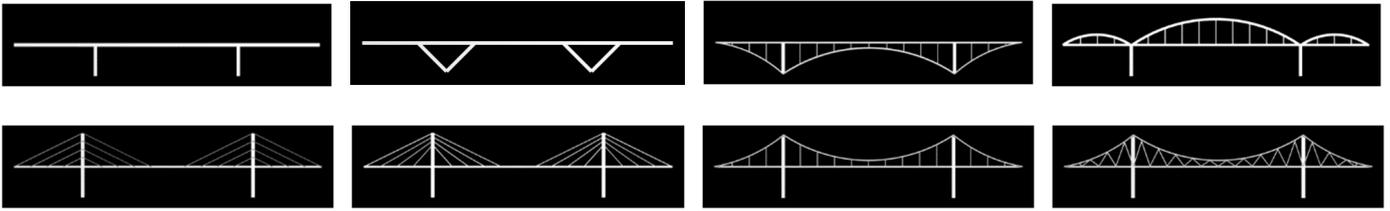

Fig. 8 Grayscale image of each bridge facade

Each sub bridge type obtained 13552 different images, resulting in a total of 108416 images in the entire dataset.

## 2.2 Construction and training of Glow

1. Constructing and training normalizing flow based on the Glow API in the TensorFlow Probability library. The specific code is as follows:

```
#https://tensorflow.google.cn/probability/api_docs/python/tfp/distributions/TransformedDistribution
flow = tfd.TransformedDistribution(
    distribution=   tfd.MultivariateNormalDiag( tf.zeros((tf.math.reduce_prod(128, 512, 1), )),
                                                tf.ones((tf.math.reduce_prod(128, 512, 1), )) ),
    bijector    =   tfb.Glow(output_shape=(128, 512, 1),
                    num_glow_blocks = 6,
                    num_steps_per_block=8,
                    coupling_bijector_fn=GlowDefaultNetwork, #self-defined
                    exit_bijector_fn=tfb.GlowDefaultExitNetwork, ), ) #default
```

Fig.9 Parameter Settings of Glow

2. The GlowDefaultNetwork module has a default number of neurons of 400, which requires too much hardware and therefore needs to be reduced. Since this parameter cannot be passed through formal parameter, the copied source code is rewritten as follows:

```
#https://tensorflow.google.cn/probability/api_docs/python/tfp/bijectors/GlowDefaultNetwork
#2024.01.07,The only modification: "num_hidden=400" to "128"
class GlowDefaultNetwork(tfk.Sequential):
    def __init__(self, input_shape, num_hidden=128, kernel_shape=3):
        this_nchan = input_shape[-1] * 2
        conv_last = functools.partial(
            tfkl.Conv2D,
            padding='same',
            kernel_initializer=tf.initializers.zeros(),
            bias_initializer=tf.initializers.zeros())
        super(GlowDefaultNetwork, self).__init__([
            tfkl.Input(shape=input_shape),
            tfkl.Conv2D(num_hidden, kernel_shape, padding='same',
                    kernel_initializer=tf.initializers.he_normal(),
                    activation='relu'),
            tfkl.Conv2D(num_hidden, 1, padding='same',
                    kernel_initializer=tf.initializers.he_normal(),
                    activation='relu'),
            conv_last(this_nchan, kernel_shape)
        ])
```

Fig.10 Parameter Settings of GlowDefaultNetwork

3. To eliminate the problem of inf and nan in loss, it is necessary to set the learning rate of optimizer Adam to 0.0001 and perform weight gradient clipping (clipnorm=1).

4. tfd.TransformedDistribution module provides "log_prob()" method to output the probability density of the sample in the sample space, which can be used by users to directly build loss function, so that users who lack advanced mathematical background can quickly build their own normalizing flow model.

　　5. Due to the constraints of hardware conditions, it is impossible to fully optimize. Thus, I can only test parameters while sampling, and generate bridge images from multiple saved generator models.

## 2.3 Exploring new bridge types through latent space sampling

　　Using "sample()" method of tfd.TransformedDistribution module to sample. Based on the thinking of engineering structure, five technically feasible new bridge types are obtained through manual screening, which were completely different from the dataset:

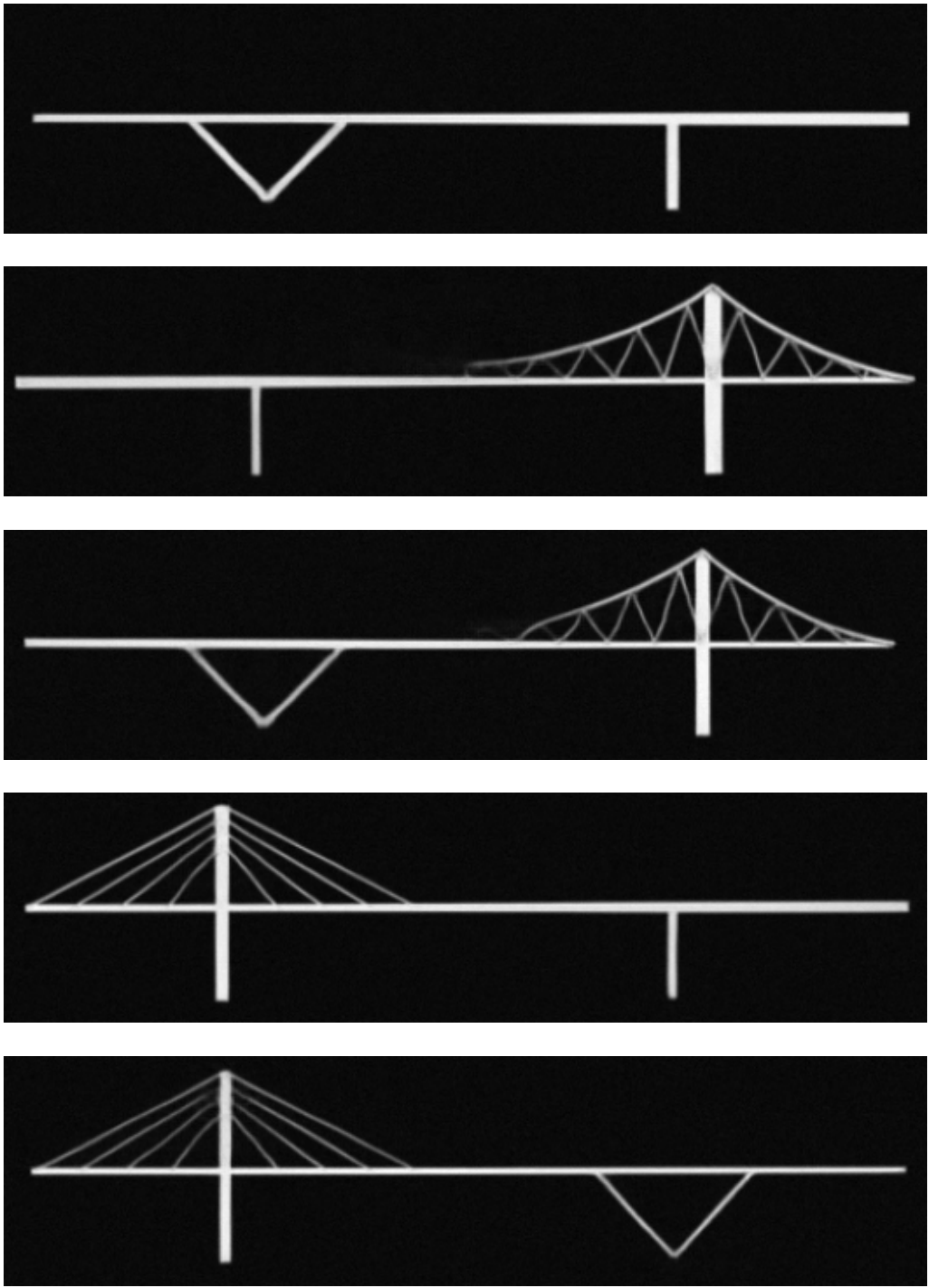

Fig.11 Five new bridge types with feasible technology

　　The new bridge type here refers to a type of bridge that has never appeared in the dataset, but is created by neural network based on algorithms, which represents the model's innovative ability.

## 2.4 Result analysis

　　The bridge types of the dataset are all symmetric structures, while normalizing flow can generate asymmetric bridge types, which are not simply superposition, but organic combinations of different structural components. It is similar to generative adversarial networks.

## 3 Conclusion

Normalizing flow is similar to generative adversarial network and is more creative than variational autoencoder. It can organically combine different structural components on the basis of human original bridge types, creating new bridge types. It has a certain degree of human original ability, and can open up the imagination space and provide inspiration to humans.